# Transforming Multidimensional Time Series into Interpretable Event Sequences for Advanced Data Mining

Xu Yan
Trine University
Phoenix, USA

Yaoting Jiang
Carnegie Mellon University
Pittsburgh, USA

Wenyi Liu
Independent Researcher
Nanjing, China

Didi Yi
Independent Research
Seattle, USA

Jianjun Wei*
Washington University in St. Louis
St Louis, USA

*Abstract*—This paper introduces a novel spatiotemporal feature representation model designed to address the limitations of traditional methods in multidimensional time series (MTS) analysis. The proposed approach converts MTS into one-dimensional sequences of spatially evolving events, preserving the complex coupling relationships between dimensions. By employing a variable-length tuple mining method, key spatiotemporal features are extracted, enhancing the interpretability and accuracy of time series analysis. Unlike conventional models, this unsupervised method does not rely on large training datasets, making it adaptable across different domains. Experimental results from motion sequence classification validate the model's superior performance in capturing intricate patterns within the data. The proposed framework has significant potential for applications across various fields, including backend services for monitoring and optimizing IT infrastructure, medical diagnosis through continuous patient monitoring and health trend analysis, and internet businesses for tracking user behavior and forecasting sales. This work offers a new theoretical foundation and technical support for advancing time series data mining and its practical applications in human behavior recognition and other domains.

I. INTRODUCTION

Time series refers to a series of observations generated by a specific variable that are recorded in sequence within a certain period of time. In the real world, many applied studies in life and industry need to analyze the collected time series data to better understand the laws of things [1]. For example, backend business processing analysis of internet companies utilizes time series data to monitor and optimize server performance, predict system load, and enhance service reliability and efficiency. Similarly, data in the field of medical diagnosis and stock data in the financial market are also forms of time series [2]. Since these types of data have a common attribute-time series, they can all be described in the form of time series. However, the effective information in the time series is usually hidden in the sequence fragments, and the unpredictability of its length, number, and shape variables adds a lot of difficulties to the mining task of time series [3]. Therefore, the analysis of time series needs to obtain information with potential value from historical data as a basis for predicting the future development trend of things, which plays a decisive role in the mining of patterns in time series [4].

The data values in the time series are arranged in chronological order. Unlike structured data, they are continuous on the time axis, and there is a logical relationship between the previous and subsequent time points (sites), which makes it difficult to directly use traditional data mining methods to carry out research[5]. For multidimensional time series, in addition to the above-mentioned temporal relationships, there are usually linear or nonlinear coupling relationships between the dimensional sequences, which manifests as a complex spatial structure of multidimensional time series[6]. The purpose of feature mining for multidimensional time series is to deeply explore the spatiotemporal structural features contained therein, so as to map the multidimensional time series data into a low-dimensional feature vector space, and provide a structured data representation model for downstream data mining tasks such as multidimensional time series clustering and classification[7,8]. This paper focuses on the problem of interpretable feature extraction of multidimensional coupling, proposes a variety of new methods for spatiotemporal feature mining of multidimensional time series, and applies them to the task of human behavior recognition based on motion sequences, which has important theoretical significance and application value.

From the three perspectives of data objects, feature representation methods and feature mining methods, the existing main time series feature mining methods are sorted out, and the advantages and disadvantages of various methods are discussed in combination with different types of data mining tasks and application scenarios, and the sequence mining process framework based on time series data representation is summarized. On this basis, the idea of efficiently mining spatiotemporal structural features based on interpretable subsequences from multidimensional time series is proposed, laying the foundation for subsequent research work.

## II. METHOD

This paper proposes a new method for representing the spatiotemporal structure of optimal prefix subsequences to efficiently extract the spatiotemporal features of motion sequences. First, a transformation method based on spatially varying events is proposed. MTS is converted into a one-dimensional event sequence, and a series of event symbols are used to represent the spatial structural information of multidimensional coupling in the sequence, which has good interpretability; secondly, a variable-length tuple mining method is proposed to extract non-redundant key event subsequences from the event sequence as the spatiotemporal structural features of the motion sequence, which is an unsupervised method that does not rely on large-scale training samples. Based on this, a new model for representing the spatiotemporal structural features of multidimensional time series is defined. Finally, the ideal performance of STEM is verified through pattern classification experiments on multiple motion sequences. The overall network architecture is shown in Figure 1

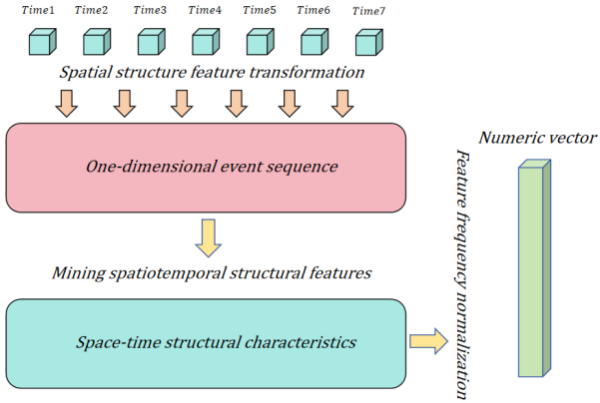

Figure 1 overall network architecture

In the multidimensional time series set $S$, each sample $X^{(i)}$ is a numerical sequence $X_{d1}^{(i)} X_{d2}^{(i)} ... X_{dT}^{(i)} \quad \forall t \in [1, T]$ in each dimension. In some specific application fields, the length of each sequence sample is different. In this case, let $T$ be the longest sequence length. For sequences shorter than $T$, the last observation value is appended (that is, it is assumed that the subsequent observation values remain unchanged). In order to analyze the coupling relationship between multidimensional sequences, this paper proposes a method of spatially changing events (SCE) to represent the structure of multidimensional sequences.

First, normalize each coordinate sequence to the interval [0,1], then track the direction of movement of each point relative to the previous point starting from the second sequence point, and detect the direction of spatial change based on the relative spatial relationship of the points. As shown in the following formula:

$$m_{dt}^{(i)} = \begin{cases} 0 & |x_{d(t-1)}^{(i)} - x_{dt}^{(i)}| \leq \delta \\ 1 & |x_{d(t-1)}^{(i)} - x_{dt}^{(i)}| > \delta \\ -1 & x_{d(t-1)}^{(i)} - x_{dt}^{(i)} < -\delta \end{cases} \quad (1)$$

Secondly, construct the sequence space change event SCE and event set E. The D-dimensional combination of symbols shown in the above formula intuitively reflects the coupling relationship between the D-dimensional sequence observations. In view of this, the SCE set is constructed by the permutation and combination of the three symbols in the formula, that is:

$$E = \otimes_{d=1}^{D} \{0, 1, -1\} \quad (2)$$

Each symbol in the sequence data represents the coupling relationship between different dimensions of the data. This representation helps to intuitively illustrate the spatial structure characteristics within the multidimensional sequence, making the data's internal relationships more understandable and interpretable. This ability to visualize and interpret the data's spatial characteristics is crucial for gaining insights into the underlying patterns and interactions present within the sequence. To analyze and extract meaningful features from these sequences, all tuples in the dataset can be organized into a collection of prefix trees, where each tree represents a different subset of the sequence data. These prefix trees are structured hierarchically, with each node representing a specific tuple and paths from the root to the leaf nodes reflecting various sequences of tuples. The Root-to-Leaf Sequence (RTS) features, which are the key features of interest, are distributed along these paths from the root to the leaf nodes in the prefix trees.

To efficiently mine RTS features, a bottom-up pruning strategy can be employed, similar to the techniques used in decision trees. This approach involves starting from the leaf nodes and progressively moving towards the root, systematically eliminating less significant nodes and paths based on certain criteria. This pruning process helps to narrow down the search space and focus on the most relevant and significant RTS features, improving the overall efficiency of the feature extraction process.

For illustration, this article provides an example of a prefix tree constructed from 3-tuples, as depicted in Figure 2. In this example, each 3-tuple serves as a node within the tree, and the tree structure visually represents the relationships and hierarchies among these tuples. By examining the paths from the root to the leaf nodes in this prefix tree, one can extract and analyze the RTS features, gaining valuable insights into the sequence data's structure and characteristics.

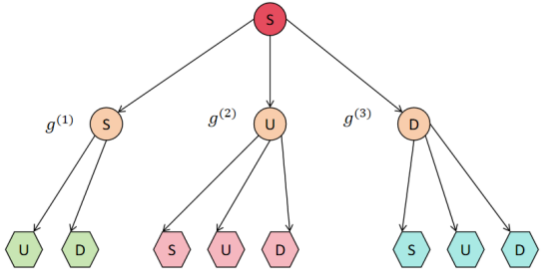

Figure 2  A prefix tree example

## III. EXPERIMENT

### A. Experimental setup

This section mainly verifies the proposed multidimensional time series spatiotemporal structure representation model. Experiments are conducted on multiple multidimensional time series datasets in practical applications and compared with several representative sequence representation models and sequence classification mining methods. All experiments are run on a personal computer with a 2.90GHz i7CPU, 16G RAM, and Windows 10 operating system.

### B. Datasets

The experiment used four public datasets from three different application fields. The first dataset, HAR (Human Activity Recognition) [9], is a human activity time series dataset. The data is collected from the embedded 3-axis accelerometer and 3-axis angular velocity meter in the smartphone device. The experiment used the 3D time series recorded by tBodyAcc_mx, tBodyAcc_ mean_y and tBodyAcc_mean_z, with lengths ranging from 9 to 47, and divided into four activity types (standing, sitting, walking, and lying). In the table, both datasets are LIBRAS gesture trajectory data, but contain different gesture types. This dataset can be obtained from the UCI Machine Learning Repository [10]. Each gesture action is represented by a 45-dimensional coordinate sequence on its motion trajectory, so it is a fixed-length 2D time series set. LIBRAS1 includes five gesture types: curved-swing, anti-clockwise-arc, vertical-swing, horizontal-swing, and clockwise-arc. The five gesture types of LIBRAS2 are vertical-zigzag, horizontal-wavy, vertical-wavy, circle, and face-up-curve.

The fourth experimental data is the WISDM dataset from the Wireless Sensor Data Mining (WISDM) laboratory [11]. Each sample is a three-dimensional time series recording human activities, corresponding to four types of activities: jogging, walking, sitting, and standing. The characteristics of this dataset are that the number of samples is small but the sequence length span is large.  This verifies the performance of the algorithm proposed in this paper on sequence sets of different numbers and lengths.

### C. Experimental Results

The performance comparison includes classification accuracy and time efficiency. Table 1 lists the average classification accuracy of different methods on each sequence set, and Table 2 summarizes the CPU time (seconds) consumed by each method.

Table 1. Comparison of average accuracy of different sequence representation methods and classifiers

| Model | HAR | LIBRAS1 | LIBRAS2 | WISDM |
|---|---|---|---|---|
| CAE | 0.57 | 0.61 | 0.63 | 0.62 |
| TFIDF | 0.59 | 0.66 | 0.66 | 0.66 |
| N-gram | 0.63 | 0.69 | 0.69 | 0.68 |
| LSTM | 0.67 | 0.74 | 0.70 | 0.76 |
| OWBC | 0.69 | 0.76 | 0.72 | 0.77 |
| TAPNET | 0.70 | 0.81 | 0.75 | 0.79 |
| Ours | 0.71 | 0.83 | 0.79 | 0.81 |

Table 2. The CPU time (in seconds) consumed by different methods to complete sequence classification

| Model | HAR | LIBRAS1 | LIBRAS2 | WISDM |
|---|---|---|---|---|
| CAE | 13.7 | 13.6 | 13.8 | 13.1 |
| TFIDF | 11.3 | 12.6 | 12.6 | 12.7 |
| N-gram | 10.2 | 11.7 | 10.6 | 11.5 |
| LSTM | 7.8 | 9.9 | 9.5 | 10.6 |
| OWBC | 6.9 | 7.8 | 8.9 | 9.6 |
| TAPNET | 5.7 | 6.7 | 8.6 | 9.3 |
| Ours | 5.2 | 6.1 | 7.9 | 8.9 |

Judging from the data in Table 1, the average accuracy rates of different sequence representation methods and classifiers on the four data sets of HAR, LIBRAS1, LIBRAS2 and WISDM are different. Our model shows the highest accuracy on all tested datasets, especially on the LIBRAS1 dataset, where the accuracy reaches 0.83, significantly better than other models. In contrast, the traditional CAE model has the weakest performance on these four datasets, with average accuracy ranging from 0.57 to 0.63, while text processing techniques such as TFIDF and N-gram perform well on some datasets. has improved, but is still not as good as the performance of deep learning models such as LSTM, OWBC and TAPNET.

Continuing to observe the data in Table 2 about the CPU time required by different models to complete sequence classification, we can find that our model not only leads in accuracy, but also performs well in computational efficiency. Specifically, our model required the least CPU time on all tested datasets, ranging from 5.2 seconds to 8.9 seconds. In comparison, a model like CAE, although not as accurate as other models, takes 13.7 seconds to complete the classification task on the HAR dataset, which is approximately twice the time required by our model. Even the relatively efficient TAPNET model takes 9.3 seconds to complete the classification task in the slowest case, which shows that our model can effectively reduce the consumption of computing resources while ensuring high accuracy.

In summary, our model shows significant advantages both from the perspective of classification accuracy and computational efficiency. It not only provides more accurate prediction results on multiple data sets, but also completes the task in a shorter time, which is crucial for practical applications, especially real-time systems. This advantage means that our model can better adapt to the needs of large data volumes and high-speed processing without sacrificing performance, thereby providing users with a more efficient service experience.

In order to better demonstrate the superiority of our algorithm, we use graphs to show our experimental results, and the display effect graphs are shown in Figures 3 and 4.

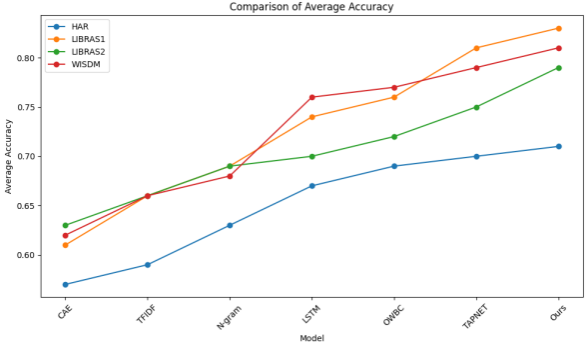

Figure3 Comparison of average accuracy

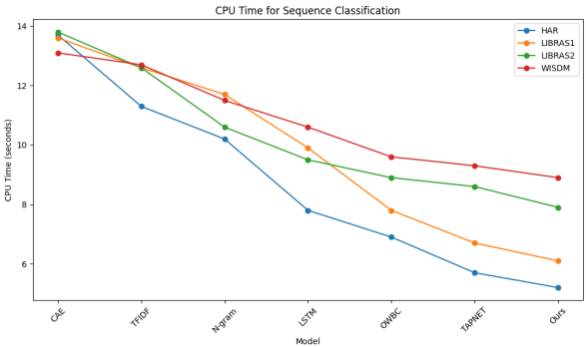

Figure 4 Comparison of time consumed

As illustrated by the figure, our newly proposed model excels in both accuracy and computational efficiency compared to existing models. It not only outperforms other models in terms of average accuracy but also demonstrates a substantial reduction in CPU time required for sequence classification tasks. This indicates that our model achieves a higher level of precision while efficiently utilizing computational resources. The reduced CPU time translates into lower energy consumption and faster processing, making it particularly advantageous for real-time applications where both speed and accuracy are essential. The dual advantage of enhanced accuracy and improved computational efficiency positions our model at the forefront of state-of-the-art solutions for sequence classification problems. This combination is especially beneficial for large-scale deployments where minimizing operational costs and maximizing performance are critical. By adopting our model, organizations can expect not only to achieve superior results but also to reduce overall energy expenses and improve response times, thereby delivering greater business value and contributing positively to societal benefits. This makes our model a highly effective choice for applications where optimizing both performance and cost is crucial.

IV. CONCLUSION

This research introduces a sophisticated spatiotemporal structure representation model specifically designed for multidimensional time series data, addressing significant limitations inherent in existing feature representation methodologies. By implementing a spatial transformation approach, this model innovatively converts complex multidimensional time series into comprehensible one-dimensional sequences of spatially evolving events. It further advances the field by introducing a novel spatiotemporal feature mining technique that extracts non-redundant, variable-length tuples as spatiotemporal features, thereby preserving the intricate original structure and richness of the data. This innovative approach effectively overcomes the typical drawbacks of models that assume independence among dimensions, providing a more nuanced analysis that significantly enhances the quality and interpretability of the data features.

Experimental validation on a motion sequence dataset has demonstrated that our model substantially improves the quality and interpretability of the extracted features compared to existing models, showcasing its superiority in handling complex datasets. The implications of this advancement are profound across several critical sectors. In backend services, for instance, the model's enhanced data processing capabilities can significantly improve system efficiencies, supporting the development of more robust solutions that are crucial for real-time, large-scale data operations. In the medical field, the model's precise and interpretable feature extraction enables the development of sophisticated diagnostic tools that can detect subtle patterns in patient data, leading to earlier and more accurate diagnoses, and ultimately improving patient outcomes. Additionally, in the realm of Internet business, this model offers valuable insights into user behavior and system performance through advanced analysis of time series data, helping businesses optimize their services and enhance user experiences, thereby fostering growth and innovation in the digital marketplace. In summary, the spatiotemporal structure representation model proposed in this paper not only marks a significant leap forward in the analytical capabilities for multidimensional time series data but also serves as a pivotal innovation with extensive applications in backend services, medical diagnosis, and internet business. Its deployment is set to revolutionize the way data-driven strategies are formulated and implemented across these sectors, enhancing the ability of industries to analyze and leverage complex datasets effectively.